RESEARCH ARTICLE

# Compact Model Parameter Extraction via Derivative-Free Optimization

RAFAEL PEREZ MARTINEZ[1], (Graduate Student Member, IEEE),
MASAYA IWAMOTO[2], (Member, IEEE), KELLY WOO[1], (Graduate Student Member, IEEE),
ZHENGLIANG BIAN[1], (Graduate Student Member, IEEE), ROBERTO TINTI[3],
STEPHEN BOYD[1], (Life Fellow, IEEE), AND
SRABANTI CHOWDHURY[1], (Fellow, IEEE)
[1]Department of Electrical Engineering, Stanford University, Stanford, CA 94305, USA
[2]Keysight Technologies Inc., Santa Rosa, CA 95403, USA
[3]Keysight Technologies Inc., Calabasas, CA 91302, USA

Corresponding author: Rafael Perez Martinez (rafapm@stanford.edu)

This work was supported in part by Stanford Graduate Fellowship (SGF) and in part by the nano@Stanford labs, part of the National Nanotechnology Coordinated Infrastructure, funded by the National Science Foundation under Award ECCS-2026822.

**ABSTRACT** In this paper, we address the problem of compact model parameter extraction to simultaneously extract tens of parameters via derivative-free optimization. Traditionally, parameter extraction is performed manually by dividing the complete set of parameters into smaller subsets, each targeting different operational regions of the device, a process that can take several days or weeks. Our approach streamlines this process by employing derivative-free optimization to identify a good parameter set that best fits the compact model without performing an exhaustive number of simulations. We further enhance the optimization process to address three critical issues in device modeling by carefully choosing a loss function that focuses on relative errors rather than absolute errors to ensure consistent performance across different orders of magnitude, prioritizes accuracy in key operational regions above a specific threshold, and reduces sensitivity to outliers. Furthermore, we utilize the concept of train-test split to assess the model fit and avoid overfitting. We demonstrate the effectiveness of our approach by successfully modeling a diamond Schottky diode with the SPICE diode model and a GaN-on-SiC HEMT with the ASM-HEMT model. For the latter, which involves extracting 35 parameters for the ASM-HEMT DC model, we identified the best set of parameters in under 6,000 trials. Additional examples using both devices are provided to demonstrate robustness to outliers, showing that an excellent fit is achieved even with over 25% of the data purposely corrupted. These examples demonstrate the practicality of our approach, highlighting the benefits of derivative-free optimization in device modeling.

## I. INTRODUCTION

Semiconductor device compact models play a crucial role in the design and development of integrated circuits and systems, serving as the bridge between physical semiconductor devices and electronic design automation (EDA) tools. These models represent mathematically the electrical (and, in some instances, thermal) behavior of semiconductor devices (i.e., charges and currents) such as transistors and diodes as a function of electrical bias. Compact models maintain a sufficient level of simplicity to be seamlessly integrated into circuit simulators, offering faster simulation times than

The associate editor coordinating the review of this manuscript and approving it for publication was Rahul A. Trivedi.

Technology Computer-Aided Design (TCAD) models while still providing the accuracy required to deliver reliable and valuable results for circuit designers. Before employing these models to design circuits and systems, it is essential to diligently extract the relevant model parameters tailored to the chosen semiconductor process, ensuring that the compact model can accurately reproduce the characteristics of a specific semiconductor device. This process involves adjusting the parameters of the compact model to align with data from the semiconductor device, whether obtained through experimental measurements or simulated using TCAD tools. However, parameter extraction has become increasingly complicated as modern compact models contain hundreds of model parameters, which are required to model the non-idealities of emerging Silicon (Si) devices such as FinFETs or III-V devices such as Gallium Nitride (GaN) high-electron-mobility transistors (HEMTs). For example, the latest version of the Berkeley Short-Channel IGFET Model (BSIM)-Common Multi-Gate (CMG) features over a thousand model parameters [1], whereas the most recent version of the Advanced SPICE Model for GaN HEMTs (ASM-HEMT) includes more than 200 model parameters [2].

Traditionally, manual fitting has been the default approach for parameter extraction in compact models. Given the extensive number of model parameters in modern compact models, a common strategy is to divide the complete parameter set into several smaller subsets [3]. These subsets usually correspond to specific physical elements of the device, such as drain/gate currents or junction capacitances. However, this approach typically involves a series of iterative steps and can extend over several days or weeks just to extract a single model card. It also frequently leaves engineers uncertain whether the resulting model card represents a near-optimal set of parameters or if there is room for further improvement.

Another common approach to reducing manual adjustment efforts is using gradient-based optimization methods. This involves employing a numerical nonlinear optimizer in conjunction with the Levenberg-Marquardt algorithm [4], [5] to extract model parameters [6]. However, obtaining gradient information in this context is exceptionally challenging. When conducting actual experiments, it is simply not possible to get gradients. Additionally, when simulations involve TCAD or SPICE, this task becomes very complicated. Gradient-based approaches are typically computationally inefficient, as calculating gradients (i.e., determining how a small change in each parameter affects the output) usually requires computationally intensive numerical approximations.

Since extracting model parameters in semiconductor devices is a complex and time-consuming task, several approaches have been proposed due to the lack of a universal method applicable across different semiconductor technologies. Among the proposed solutions, a notable deep learning approach involves training a neural network to output the desired model parameters using the device's characteristics as inputs [7], [8], [9], [11], [12]. The device characteristics (or inputs) may include data that is either measured or simulated using TCAD, encompassing $I-V$ characteristics, S-parameters, or large-signal load-pull data. While this approach is generally practical, it demands substantial computational resources, requiring thousands of simulations to train the neural network just to extract the parameters of one type of device. This becomes particularly inefficient when fitting a new type of semiconductor device each time. The complexity of this approach is further increased by the necessity of prior parameter extraction experience to define the variation ranges of the model parameters, which is essential for generating training data for the desired device.

We propose adopting derivative-free optimization (DFO) to address the issues above. DFO methods approximately minimize a function only using the objective value (i.e., no gradients are required). They are also straightforward to implement and more computationally efficient. The primary advantage of our approach lies in its ability to identify a set of model parameters that achieves a near-optimal fit with significantly fewer simulations than would be required for a full-grid search. This approach not only mitigates the curse of dimensionality often encountered when considering tens of parameters but also allows us to explore a relatively wide range of plausible parameter values.

To further enhance the optimization process, we carefully select a loss function that addresses three key issues: 1) ensuring consistent model performance across different orders of magnitude by focusing on relative errors rather than absolute errors; 2) guiding the optimization process to prioritize regions of particular interest while deprioritizing less critical regions of operation; and 3) reducing sensitivity to outliers and measurement errors. Moreover, we utilize a standard model assessment method (train/test split) used in Statistics and machine learning (ML) to judge the fit of our extracted model [13]. This method is unlike traditional approaches in device modeling that fits the model to the entire dataset, potentially leading to overfitting.

The remaining paper is structured as follows: Section II formulates the problem we are trying to solve (i.e., model parameter extraction) as an optimization problem in a more general form and introduces the DFO framework we used in the present work. Section III describes our proposed approach and the loss function we chose to tackle this problem. It also includes a straightforward example of fitting a simple two-terminal device, specifically a diamond Schottky diode, to clearly outline and effectively demonstrate the issues we are addressing. Section IV presents the modeling of a 150-nm gate length ($L_G$) GaN-on-SiC HEMT using the ASM-HEMT DC model, a task that involves extracting more than 30 model parameters simultaneously. Lastly, Section V concludes this article.

## II. DFO FOR MODEL PARAMETER EXTRACTION
In this work, we focus on extracting semiconductor compact model parameters. This process involves identifying a set of parameter values that precisely replicate a device's characteristics, which may derive from various experiments on a fabricated device or TCAD simulations. These characteristics can be static (e.g., $I-V$ characteristics or S-parameters), dynamic (e.g., dynamic load-lines), or a combination of both (i.e., heterogeneous).

In the context of the present work, each experiment involves collecting one or more measurements from the device. Considering $k$ measurements, each $y_i$ corresponds to a distinct measurement obtained from an experiment. For example, in the case of a diode, a single experiment might involve measuring multiple points on the $I$-$V$ curve, where each measurement captures the current $I_i$ for a given voltage $V_i$. Collectively, these measurements represent the $I$-$V$ characteristics of the diode. The output of the compact model is denoted by $\hat{y}_i$ and parameterized by a $p$-vector $\theta = (\theta_1, \ldots, \theta_p)$, where $\theta$ is a vector of length $p$ (i.e., $\theta$ is a vector with $p$ parameters), and lies within a subset $\Theta \subseteq \mathbb{R}^p$ that represents the selected model parameters from a feasible set. Each model parameter in $\theta$ has a range of plausible values. In some cases, it is more convenient to work with the logarithm of the parameter's range (e.g., the saturation current in a diode). We seek to solve the optimization problem:

$$\begin{aligned} \text{minimize} \quad & \frac{1}{k}\sum_{i=1}^{k}\mathcal{L}(\hat{y}_i, y_i) \\ \text{subject to} \quad & \theta \in \Theta, \end{aligned} \quad (1)$$

where $\mathcal{L}(\hat{y}_i, y_i)$ is the loss function we want to minimize over the model parameters $\theta$. Here, $\theta$ represents the variable in the optimization problem, $\hat{y}_i$ is the predicted or simulated value obtained from the compact model, and $y_i$ is the true value that has been experimentally measured or simulated using TCAD.

The complexity of solving this optimization problem is that the model's output is not given by simple expressions but rather by running a SPICE simulation. This implies that obtaining gradient information is generally challenging and prohibitively expensive. This difficulty is compounded when considering tens of model parameters, which turns this task into a computationally intensive endeavor. This is a classic case of the curse of dimensionality, where exploring every potential combination of parameter values becomes impractically costly in terms of time and computational resources. In such scenarios, DFO excels at finding near-optimal solutions as this approach does not require gradients and directs the optimization effort towards promising zones of the parameter space that are more likely to yield the most accurate model fits. This significantly reduces the computational overhead by eliminating the need for calculating gradients and decreasing the number of simulations required to obtain a good fit [14].

DFO methods have also shown promise in ML, particularly in hyperparameter tuning. In the context of ML, hyperparameters are settings or configurations that control the behavior of a machine learning algorithm, such as the learning rate or the number of hidden layers in a neural network. DFO methods are effective at identifying a good set of hyperparameters that significantly enhance ML model performance [15], [16], [17], [18].

Two drawbacks of DFO methods are their reduced effectiveness when dealing with hundreds of parameters and their inability to guarantee the attainment of a global solution [19]. This is primarily due to the curse of dimensionality, which makes it challenging to sample points close to the global optimum unless a large number of samples are taken. Additionally, DFO methods employ stochastic or heuristic sampling methods to explore the design space, which could lead to settling on local optima without guaranteeing a global solution. Nevertheless, the problem we are addressing is in the order of tens of parameters and results in a nearly optimal fit with far fewer simulations than what would be required by a full-grid search. This represents a clear advantage over the deep learning approaches used in previous works by [7], [8], [9], [10], and [12], which require thousands of simulations to generate their training and test datasets.

It is also worth noting that the problem we are trying to solve is analogous to calibrating a TCAD model, which involves matching experimental data with the simulated results from a TCAD device simulator by adjusting TCAD model parameters. Given that TCAD simulations are generally more expensive and time-consuming, our proposed approach is particularly advantageous. In such scenarios, DFO methods excel at finding a nearly optimal fit with far fewer simulations than a full-grid search, which is a clear advantage over manually tuning TCAD model parameters, which can take weeks or, in some instances, months to achieve a good fit.

### A. DERIVATIVE-FREE OPTIMIZATION FRAMEWORK

We resort to open-source hyperparameter optimization frameworks in the present work since they employ DFO methods. Some of them include Autotune [15], HOLA [16], Hyperopt [17], and Optuna [18], to name a few. A primary advantage of these frameworks is their streamlined application programming interface (API), which enables users to configure a parameter search space with minimal coding effort. This allows for quick adjustments to the loss function or to constraints related to model parameters. The choice of optimization framework is entirely up to the user, as each framework is relatively similar and shares common features. In this work, we employed Optuna and its default sampler, the Tree-structured Parzen Estimator (TPE), for model fitting due to its capability to efficiently explore large parameter spaces. Under these settings, Optuna has demonstrated effectiveness in prior studies, navigating spaces that encompass up to 34 parameters [18], which aligns with the complexity of our task.

Most DFO frameworks follow a structured approach to improve model performance through parameter tuning.

Initially, they use a sampling method to explore the parameter space and identify promising regions for optimal settings. As the process advances, these frameworks adopt refined sampling strategies, focusing on the top 20-30% of parameters that have shown the best performance. This strategic refinement enables the algorithm to gradually learn and adapt to the distribution of the most effective parameter values. As more data points are accumulated, the accuracy of identifying a superior set of parameters increases, resulting in enhanced model performance. In particular, Optuna allows the user to choose among various sampling strategies, with the default option being the TPE sampler. This sampling approach uses past outcomes to predict which parameter settings might lead to better outcomes by employing two Gaussian Mixture Models: one looks at a set of parameter values with the best results and the other looks at the remaining parameters. The TPE algorithm then decides whether to try new and untested parameter values or use the ones already shown to work well. By balancing the search for new parameters with known effective values, this strategy enhances the likelihood of finding the best set of parameters [18]. One of the limitations of the TPE algorithm is that its performance can be influenced by the quality of the initial random sampling used to initialize the algorithm [20]. Therefore, it is suggested performing multiple independent TPE runs. Despite this limitation, the algorithm consistently delivers strong performance across various independent runs. Interested readers may refer to [20] for a comprehensive tutorial on TPE.

## III. PROPOSED APPROACH FOR PARAMETER EXTRACTION

In our proposed approach, we begin by setting the device dimensions, if necessary. Following this initial step, we define the loss functions for one or more experiments along with any choice of hyperparameters. We also set a plausible wide range of values for each model parameter, with some ranges in the log scale, if appropriate. Subsequently, we partition the data of each experiment, allocating 80% for training and the remaining 20% for testing. After completing these steps, we begin the optimization process by using Optuna as the hyperparameter optimization (HPO) framework, along with the TPE sampling strategy. Within the HPO framework, a set of parameters is sent to a simulator (in the context of this work, that would be Keysight Device Modeling IC-CAP [22]). The loss function is evaluated based on the simulator's output for the provided set of model parameters, and an error is returned. Using this information, the HPO framework makes an informed decision to choose the parameters of the subsequent trial. This procedure is repeated for a predetermined number of trials. The model extraction is then judged using the data allocated for testing. If the performance is not satisfactory, we can try a different choice of model, loss function, or hyperparameters to check if any of these settings are more suitable than others. Once we have established the choice of model, loss function, and hyperparameters, we retrain on the entire dataset. This time, however, we further tighten the upper and lower limits on the parameter ranges to be closer to the best set of model parameters identified during the train/test split. We also initiate the optimization process using the best set of parameters we have previously found. Re-training generally requires fewer trials since we already have prior knowledge of where a good set of model parameters can be found. The reasoning behind re-training on the complete dataset is to maximize the model's exposure to all available data. This allows the model to learn from the full spectrum of device behaviors and characteristics, which were withheld during the testing phase. It is especially beneficial in semiconductor modeling, where data may be limited, as it can significantly enhance the model's accuracy. This step can also be done when data from a new experiment becomes available from another device fabrication run (which can alter the device characteristics due to process variations) or when the foundry improves the device performance. The steps of our extraction process are summarized in Fig. 1(a). Furthermore, our proposed approach is compared against the conventional extraction flow of the ASM-HEMT model, as shown in Fig. 1(b), which consists of several manual iterative steps, as outlined in [21].

### A. SELECTION OF LOSS FUNCTION
In device modeling, selecting the appropriate loss function is critical to obtaining a reasonable percent error over the range of interest, ensuring a good model fit. Our proposed loss function is chosen carefully to address three significant issues in device modeling. We start with the absolute error (or L1 loss) function as a foundation and then incrementally enhance it to tackle these issues comprehensively, as given by:

$$\mathcal{L}_1(\hat{y}_i, y_i) = |\hat{y}_i - y_i|. \qquad (2)$$

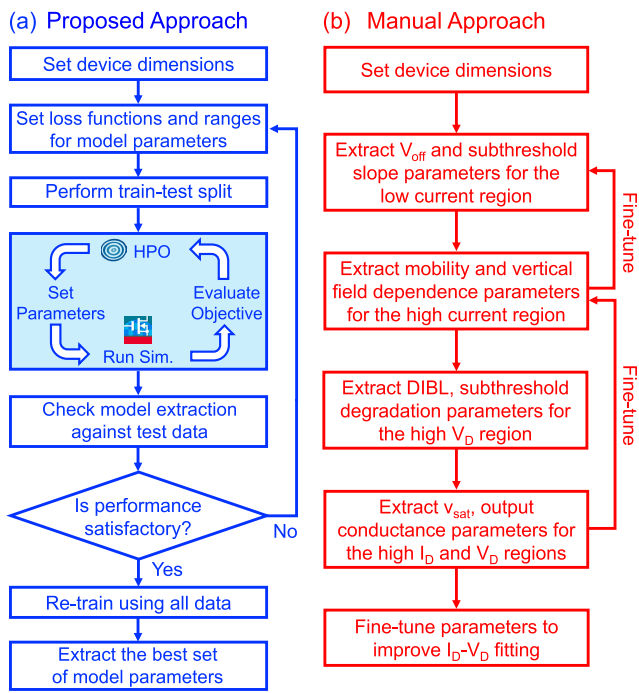

**FIGURE 1.** (a) Proposed approach to extract DC model parameters using our derivative-free optimization framework, and (b) Manual approach for extracting DC model parameters in the ASM-HEMT model (after [21]).

First, we want to ensure consistent model performance across different orders of magnitude. For example, we aim to fit the model across a wide range of current values, ranging from small to large (e.g., 10 $\mu$A to 100 mA). To achieve this, we perform a log transformation, shifting the focus of the loss function from absolute errors to relative errors. This provides a uniform assessment across different scales of data, resulting in the following equation:

$$\mathcal{L}_{\log}(\hat{y}_i, y_i) = |\log(\hat{y}_i) - \log(y_i)|. \qquad (3)$$

Second, we are interested in fitting our model in key operational regions of the device. This implies that we are not interested in fitting the model in less critical regions below some threshold, as spending effort to fit a model to irrelevant values or below the experimental or simulation noise floor will lead to a much worse model. To effectively address this, we transform the loss function once more as follows:

$$u(\hat{y}_i, y_i) = \left| \log\left(1 + \frac{\hat{y}_i}{\epsilon_i}\right) - \log\left(1 + \frac{y_i}{\epsilon_i}\right) \right|. \qquad (4)$$

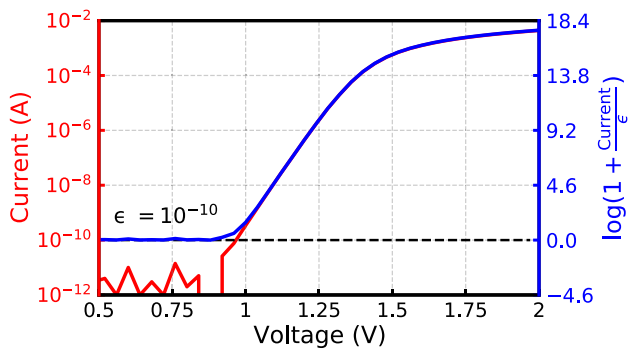

**FIGURE 2.** Transformation of the loss function that effectively excludes any current values below a threshold $\epsilon$ from the fitting process. The left y-axis (red) represents the current values without the transformation, whereas the right y-axis (blue) represents the current values after the transformation.

In this second transformation, the hyperparameter $\epsilon_i$ denotes a threshold below which the values of $\hat{y}$ and $y$ are considered negligible. This can be exemplified in Fig. 2 by considering the $I - V$ characteristics of a diamond Schottky diode. In Fig. 2, the current values on the left y-axis are represented linearly without the transformation, whereas the right y-axis represents the current values after the transformation. We observe that below the threshold $\epsilon_i$, any current values below $10^{-10}$ A are effectively excluded from the fitting process (i.e., their contribution is close to zero).

Lastly, we would like a fitting method that is robust to outliers and measurement errors. This is because, in both measurements and simulations, there are some experiments that are simply corrupted, which can occur from hitting the instrument compliance at random or when poor accuracy is obtained at higher frequencies, such as in S-parameter measurements. If proper care is not taken, outliers and measurement errors can significantly destroy the quality of the fit. Preferably, it is desired to identify potential outliers and reduce their impact on the fitting process rather than completely excluding them. This can be addressed by incorporating a clipped (non-convex) penalty function on top of (4), which is represented mathematically as

$$\mathcal{L}_{\text{clip}}(\hat{y}_i, y_i) = \begin{cases} u^2 & \text{if } |u| \leq \delta_i, \\ \delta_i^2 & \text{if } |u| > \delta_i. \end{cases} \qquad (5)$$

Here, $u$ is the transformed error term from (4), which addresses the previously discussed issues. The hyperparameter $\delta_i$ is a predefined threshold that sets the maximum allowable error. This implies that the penalty function puts a fixed cap on $|u|$ larger than $\delta_i$, regardless of size, i.e., we disregard any $|u|$ exceeding $\delta_i$, treating them as outliers or flawed data [23]. Furthermore, to maintain consistency with this methodology, we select constant values for the hyperparameters $\epsilon_i$ and $\delta_i$ for each experiment. Consequently, unless stated otherwise, the loss function defined in (5) will be the primary loss function used for fitting throughout the remainder of this work. It incorporates the enhancements and transformations previously detailed to ensure robust and accurate model fitting across various regions of operation.

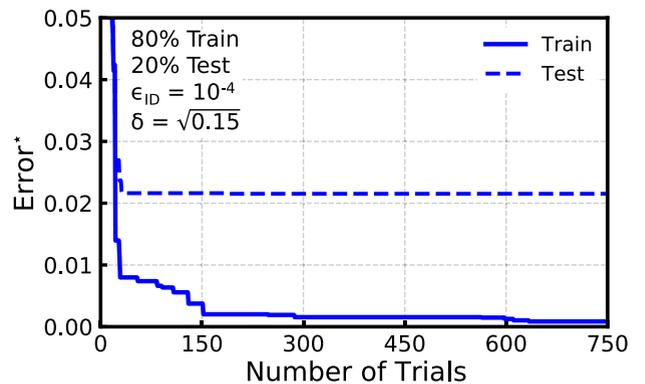

**FIGURE 3.** Error* progression across 750 trials, illustrating train and test curves. In this example, 35 model parameters were adjusted to fit 35 measurements, with 80% of the data allocated for training and the remaining for testing.

### B. IMPORTANCE OF MODEL ASSESSMENT
When performing model parameter extraction with tens of model parameters, the concept of model assessment becomes essential to ensure the efficacy of our extracted model. This principle is straightforward: the compact model should perform reliably within SPICE simulations under various voltages, currents, and conditions that may differ from our original measurements used to extract the model. We aim for the model to excel in scenarios it has not previously encountered. This is a common problem in Statistics and ML, where the objective is not to match the data used to train these models but to accurately predict new, unseen data [13].

A simple example illustrating the importance of model assessment can be seen in scenarios where we have 35 model parameters that need to be adjusted based only on 35 measurements. Here, we will use the ASM-HEMT model and 35 $I_D - V_D$ measurements of a 150-nm gate length GaN-on-SiC HEMT, selected randomly. We then split the data into 80% training and the remaining 20% for testing. The results show that the model does an excellent job fitting the training data as the minimum error observed to date (Error*) keeps decreasing as the number of trials increases. However, when evaluated against the testing data, the Error* reaches a plateau relatively early. This indicates that the model, while improving on the training dataset with more trials, shows little to no improvement on the testing dataset after 30 trials. These results are summarized in Fig. 3.

For these reasons, we adopted the concept of model assessment to judge our model's fit. In all of the fitting examples presented later in the text, we have performed this approach by splitting the data from each experiment into 80% for training and the remaining 20% for testing. The results also show that in most cases, the test curve for the minimum error observed to date is lower than the training curve. As such, this implies that our model generalizes well even for data that it has not seen.

## C. DIAMOND SCHOTTKY DIODE FITTING EXAMPLE

Having established the parameter extraction approach and the proposed loss function, we initially focus on applying these methods to a simple two-terminal device, specifically a diamond Schottky diode, using the SPICE Diode Model [24]. This demonstration aims to illustrate the effectiveness of our approach in a straightforward setting before tackling the complexities of a more sophisticated compact model. The two-terminal device we are considering is a diamond pseudo-vertical Schottky barrier diode with a diameter of 100 $\mu$m. Interested readers may refer to [25] for a more in-depth discussion on the characteristics and performance of these devices, where additional insights into diamond Schottky diodes are presented. The diamond Schottky diode was fabricated on a single crystalline diamond structure consisting of MPCVD-grown p-type epilayers on a type Ib (100) diamond substrate. Additional details of the fabrication and structure of the Schottky diode can be found in [26], where it is referred to as Sample A. Additionally, Mo was used as the Schottky contact metal for the diode measured in this study.

The SPICE diode model is fitted by adjusting three parameters: $n$ (ideality factor), IS (saturation current), and RS (series resistance). The parameter ranges for the SPICE diode model used during the train/test split are provided in Table 1. The model was trained with hyperparameters set to $\epsilon = 10^{-10}$

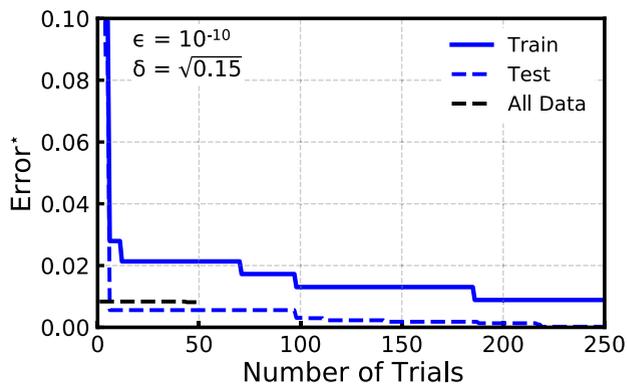

**FIGURE 4.** Error* progression across 250 trials, illustrating the train and test curves, along with the error curve when retraining on the entire dataset, in the modeling of the diamond Schottky diode. The loss function employed hyperparameter values of $\epsilon = 10^{-10}$ and $\delta = \sqrt{0.15}$.

**TABLE 1.** Parameter ranges in the SPICE diode model.

| Parameter | Lower Limit | Upper Limit |
|---|---|---|
| IS* | $1 \times 10^{-25}$ | $1 \times 10^{-22}$ |
| $n$ | 0.5 | 1.5 |
| RS | 100 | 150 |

*Log-spacing

and $\delta = \sqrt{0.15}$. Having established the loss function and the hyperparameter values, the model was then re-trained for 50 trials to further adjust the model parameters on the entire dataset, starting from the best parameters found during the train/test split. The minimum error observed to date is plotted in Fig. 4 as a function of the number of trials for both train and test curves, along with the error curve when retraining on the entire dataset. The $I - V$ characteristics of the simulated data against measurements are displayed in Fig. 5. The fit was then evaluated by considering all current values $> 10^{-10}$ A using the loss function defined in (5), yielding an error of 0.010.

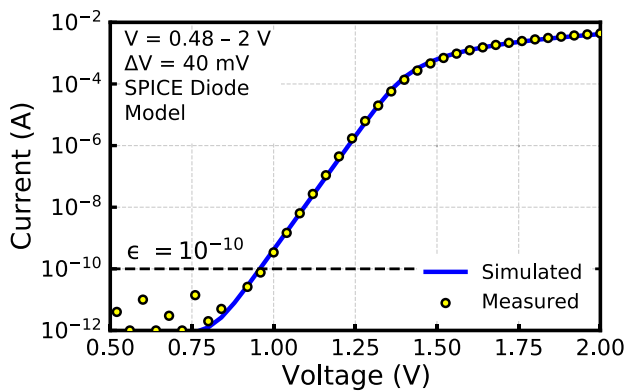

**FIGURE 5.** Measured and simulated $I - V$ characteristics of a diamond Schottky diode. The voltage sweep ranged from 0.48 to 2 V in 40 mV steps, resulting in 39 measurements.

To showcase the efficacy of our proposed loss function against measurement outliers, we intentionally corrupt 50% of the diode's measurements (in practice, this would not occur, but our goal with this example is to show that our fitting is resilient to measurement anomalies). We consider the loss function defined in (5), which is robust to outliers, as well as a separate non-robust loss function given by

$$\mathcal{L}_{2,u}(\hat{y}_i, y_i) = \left| \log\left(1 + \frac{\hat{y}_i}{\epsilon_i}\right) - \log\left(1 + \frac{y_i}{\epsilon_i}\right) \right|^2. \quad (6)$$

This loss function is similar to (5), but without the addition of the penalty function. Both loss functions are then used to repeat the model extraction procedure with the same hyperparameter values and settings as the previous diode example without data corruption. The minimum error observed to date is plotted in Fig. 6(a) as a function of the number of trials for both train and test curves, along with the error curve when re-training on the entire dataset, for the robust loss function given by (5). Similarly, Fig. 6(b) showcases the train and test curves, along with the error curve when re-training on the entire dataset, for the non-robust function without the penalty function as defined in (6). The $I-V$ characteristics of the measured (corrupted) and simulated data using the robust and non-robust loss functions are displayed in Fig. 7. Despite the data corruption, the robust loss function accurately fitted the model to the measurements, showing an error of 0.011 for currents above $10^{-10}$ A when compared against the non-corrupted data. Conversely, using the non-robust loss function resulted in a poor fit since it was heavily affected by outliers, yielding an error of 0.756 for currents $> 10^{-10}$ A when compared against the non-corrupted data.

## IV. 150-nm GaN-ON-SiC HEMT FITTING EXAMPLE

After successfully modeling a diode using our proposed approach for parameter extraction, we then shifted our focus to fitting measured DC data of a GaN-on-SiC HEMT to the ASM-HEMT DC model [2]. For an in-depth description of the broader implications and technical details of RF GaN HEMT devices and their significance in 5G and beyond-5G wireless communications, interested readers may refer to [27]. The GaN-on-SiC HEMT device we are modeling features a 150-nm gate length and a gate width ($W_G$) of $4 \times 50$ μm. We have previously reported other modeling approaches utilizing the same GaN process in [28] and [29]. In this example, we focus on the ASM-HEMT model, an industry-standard compact model for GaN HEMTs.

The ASM-HEMT model effectively captures a range of device non-idealities, including self-heating and temperature dependence, mobility degradation, drain-induced-barrier-lowering (DIBL), velocity saturation, and trapping effects [2]. The parameter extraction process for this model begins with extracting the DC model parameters, which are fundamental for determining the device's operation as a whole. Considering its importance, it is crucial to pay close attention during this initial stage to obtain an overall good model fitting. The conventional approach to extracting the ASM-HEMT DC model (as described in [21] and [30]) considers splitting the set of DC model parameters into smaller subsets to fit certain regions of operation in the $I-V$ plane through a series of iterative steps. On the other hand, our proposed DFO approach allows us to simultaneously consider 35 relevant parameters in the DC model across a relatively wide range of plausible parameter values and fit them to measured $I-V$ characteristics in a straightforward manner.

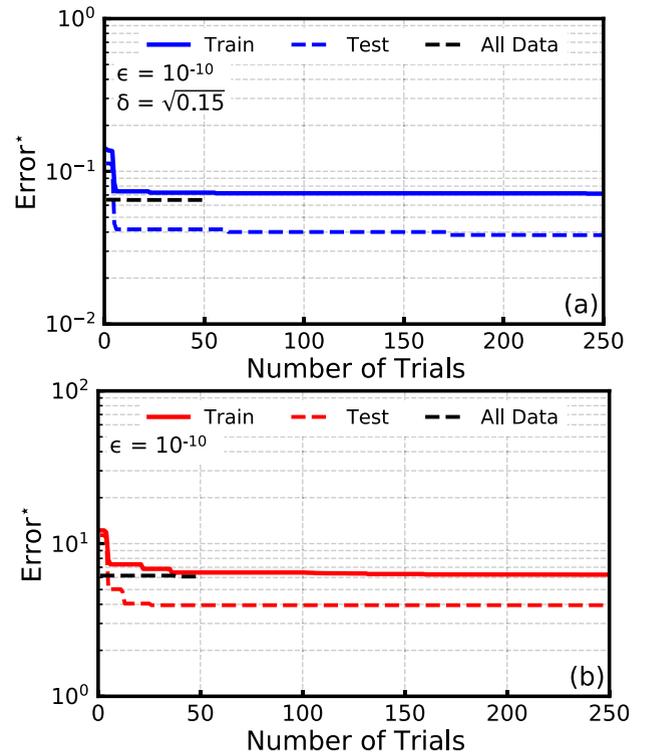

**FIGURE 6.** Error* progression over 250 trials, illustrating the train and test curves, along with the error curve when retraining on the entire dataset, in the modeling of the diamond Schottky diode for the (a) robust and (b) non-robust loss functions, with 50% of the data being corrupted. A value of $\epsilon = 10^{-10}$ was used for both loss functions, while a value of $\delta = \sqrt{0.15}$ was employed for the robust loss function.

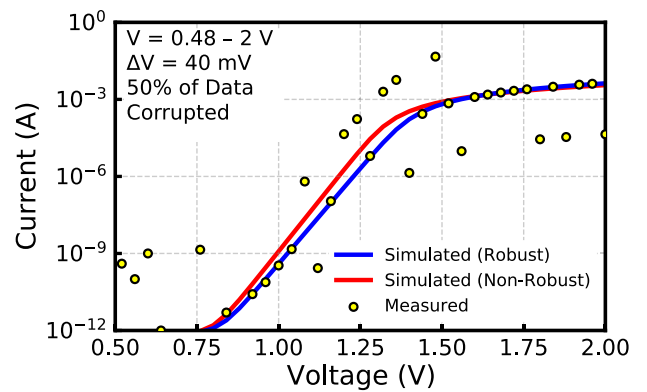

**FIGURE 7.** Measured and simulated $I-V$ characteristics of a diamond Schottky diode using both robust and non-robust loss functions. The voltage sweep ranged from 0.48 to 2 V in 40 mV steps, totaling 39 measurements. Here, 50% of the measurements were randomly corrupted to evaluate the performance of the loss functions.

**TABLE 2.** Parameter ranges in the ASM-HEMT model.

| Parameter | Lower Limit | Upper Limit |
|---|---|---|
| CDSCD | $1 \times 10^{-3}$ | $150 \times 10^{-3}$ |
| ETA0 | $10 \times 10^{-3}$ | $100 \times 10^{-3}$ |
| DELTA | 2 | 5 |
| GDSMIN* | $1 \times 10^{-12}$ | $1 \times 10^{-6}$ |
| IGDDIO | 7.5 | 15 |
| IGSDIO | 2.5 | 10 |
| IMIN* | $1 \times 10^{-15}$ | $1 \times 10^{-12}$ |
| LAMBDA | $100 \times 10^{-6}$ | $1500 \times 10^{-6}$ |
| MEXPACCS | 1 | 5 |
| MEXPACCD | 1 | 5 |
| NFACTOR | 0.2 | 0.5 |
| NJGD | 2.5 | 20 |
| NJGS | 2.5 | 20 |
| NS0ACCD* | $5 \times 10^{15}$ | $5 \times 10^{20}$ |
| NS0ACCS* | $5 \times 10^{15}$ | $5 \times 10^{20}$ |
| RDC | $100 \times 10^{-6}$ | $1500 \times 10^{-6}$ |
| RIGDDIO | $10 \times 10^{-9}$ | $100 \times 10^{-9}$ |
| RIGSDIO | $10 \times 10^{-9}$ | $100 \times 10^{-9}$ |
| RNJGD | 15 | 30 |
| RNJGS | 5 | 15 |
| RSC | $100 \times 10^{-6}$ | $1500 \times 10^{-6}$ |
| RTH0 | 31.5 | 32.5 |
| THESAT | 1 | 4 |
| U0 | $150 \times 10^{-3}$ | $300 \times 10^{-3}$ |
| U0ACCD | $50 \times 10^{-3}$ | $250 \times 10^{-3}$ |
| U0ACCS | $50 \times 10^{-3}$ | $250 \times 10^{-3}$ |
| UA | $1 \times 10^{-8}$ | $50 \times 10^{-8}$ |
| UB* | $1 \times 10^{-21}$ | $1 \times 10^{-18}$ |
| UTE | $-1$ | $-0.1$ |
| UTED | $-17.5$ | $-5$ |
| UTES | $-17.5$ | $-5$ |
| VDSCALE | 2 | 6 |
| VOFF | $-2.1$ | $-1.9$ |
| VSAT | $150 \times 10^3$ | $250 \times 10^3$ |
| VSATACCS | $10 \times 10^3$ | $150 \times 10^3$ |

*Log-spacing

In the present work, we used version 101.4.0 of the ASM-HEMT model [21]. To account for the reverse gate leakage current in GaN HEMTs caused by the Poole-Frenkel effect, we set the GATEMOD parameter to 2. This setting enables the model formulations to represent the Poole-Frenkel reverse current. Moreover, the self-heating model was incorporated, whereas the trapping and field-plate models were not. For the device's resistance characteristics, the bias-dependent access region resistance and source contact resistance were modeled by setting RDSMOD to 1. A sophisticated gate resistance model was also enabled by setting RGATEMOD to 2 [21].

For our optimization process, we selected 35 model parameters to model the DC characteristics of a GaN-on-SiC HEMT. These 35 model parameters are crucial to modeling the GaN HEMT's DC characteristics. It includes the basic core model parameters, which consist of 14 parameters: VOFF (cut-off voltage), U0 (low-field mobility), UA (mobility degradation coefficient), UB (second-order mobility degradation coefficient), VSAT (saturation velocity), DELTA (effective drain voltage exponent), LAMBDA (channel length modulation coefficient), ETA0 (DIBL parameter), VDSCALE (DIBL scaling drain-source voltage), THESAT (velocity saturation parameter), NFACTOR (subthreshold slope parameter), CDSCD (subthreshold slope change due to drain voltage), IMIN (minimum drain current), and GDSMIN (shunt conductance across the channel and field plates).

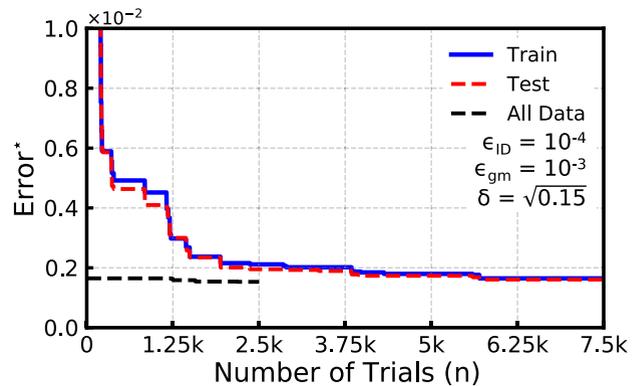

**FIGURE 8.** Error* progression across 7,500 trials, illustrating the train and test curves, along with the error curve when retraining on the entire dataset, in the modeling of the GaN-on-SiC HEMT. The loss function utilized hyperparameter values of $\epsilon = 10^{-4}$ for $I_D$ and $\epsilon = 10^{-3}$ for $g_m$, with a value of $\delta = \sqrt{0.15}$ used for both $I_D$ and $g_m$.

The parameters of the access region resistance model were also taken into account. In GaN HEMTs, access region resistances are significant because the distances between the gate edge and the source and drain edges can be a few micrometers, which can be modeled as a bias-dependent resistance [21]. Specifically, the access region resistance model parameters include VSATACCS (access region saturation velocity), NS0ACCD (2DEG density in the drain access region), NS0ACCS (2DEG density in the source access region), U0ACCD (drain side access region mobility), U0ACCS (source side access region mobility), MEXPACCD (drain access region resistance exponent), MEXPACCS (source access region resistance exponent), RDC (drain contact resistance), and RSC (source contact resistance).

Self-heating parameters were included to account for self-heating effects in the transistor, as observed in $I - V$ characteristics. These parameters consist of RTH0 (thermal resistance), UTE (temperature dependence of mobility), UTED (drain access region mobility temperature dependence), and UTES (source access region mobility temperature dependence).

Lastly, the gate current model parameters were included as they directly affect the reverse leakage current of the transistor, specifically, IGDDIO (gate-drain junction diode saturation current), IGSDIO (gate-source junction diode saturation current), NJGD (gate-drain junction diode current ideality factor), NJGS (gate-source junction diode current ideality factor), RIGDDIO (gate-drain junction reverse diode current multiplier), RIGSDIO (gate-source junction reverse diode current multiplier), RNJGD (reverse bias slope factor of the gate-drain junction diode current), and RNJGS (reverse bias slope factor of the gate-source junction diode current).

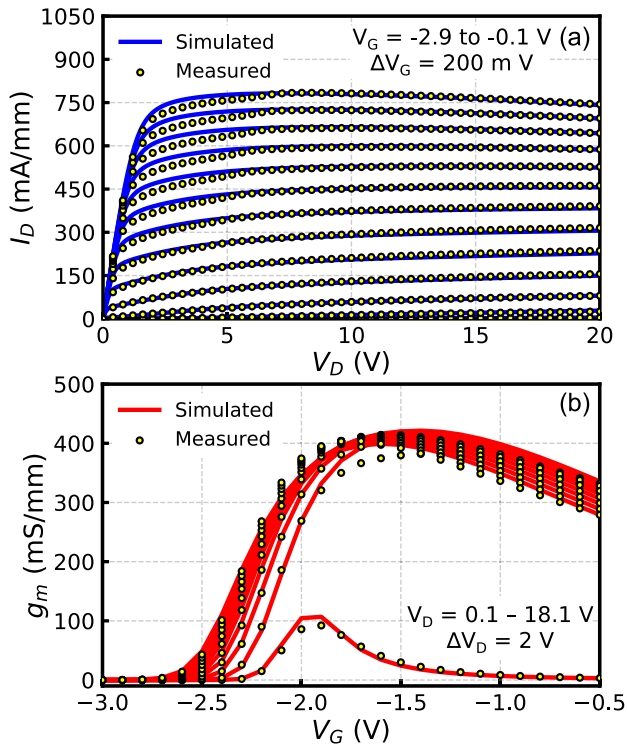

FIGURE 9. Measured and simulated (a) $I_D$ and (b) $g_m$ characteristics of the GaN HEMT. For (a), the drain voltage $V_D$ was swept from 0 to 20 V with a step size of 100 mV, and $V_G$ was swept from −2.9 to −0.1 V with a step size of 200 mV. For (b), the gate voltage $V_G$ was swept from −3 to −0.5 V with a step size of 100 mV, and the drain voltage $V_D$ was swept from 0.1 to 18.1 V with a step size of 2 V.

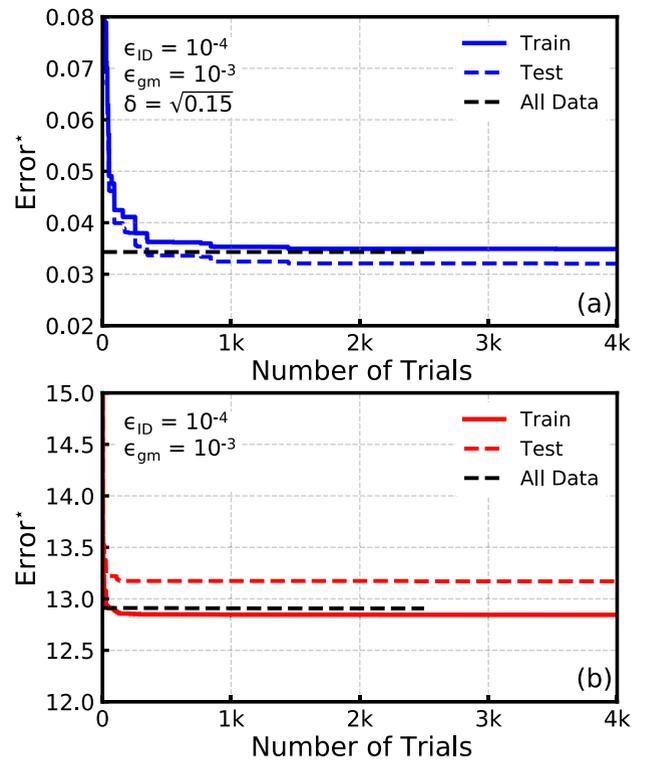

FIGURE 10. Error* progression across 4,000 trials, illustrating the train and test curves, along with the error curve when retraining on the entire dataset, in the modeling of the GaN-on-SiC HEMT for the (a) robust and (b) non-robust loss functions, with 25% of the data being corrupted. Both loss functions utilized hyperparameter values of $\epsilon = 10^{-4}$ for $I_D$ and $\epsilon = 10^{-3}$ for $g_m$, while a value of $\delta = \sqrt{0.15}$ was employed for the robust loss function in both $I_D$ and $g_m$.

We followed the steps outlined in our proposed approach, as shown in Fig. 1(a), to fit the ASM-HEMT DC model to the measured device's drain current ($I_D$) and transconductance ($g_m$) characteristics as functions of the gate voltage ($V_G$) and drain voltage ($V_D$). The voltages of the two experiments were swept from 0 to 20 V ($\Delta V_D = 100$ mV) and −3 to −0.1 V ($\Delta V_G = 100$ mV) for $V_D$ and $V_G$, respectively. This results in a total of 6,030 measurements for $I_D$ and 6,030 measurements for $g_m$. The range of the model parameters is chosen to cover a wide range of plausible values, except RTH0, which was extracted utilizing the technique outlined in [31], and VOFF, which can be extracted with relative ease. The parameter ranges for the ASM- HEMT model used during the train/test split are provided in Table 2. Given that we are dealing with a multi-objective optimization problem, as it involves two experiments (i.e., $I_D$ and $g_m$), we apply a simple scalarizer to transform the multi-objective problem into a single-objective problem. This weighted sum scalarizer is given by

$$\phi(u) = w^T v, \quad (7)$$

where $w \in \mathbb{R}^n$ is a set of weights, with $w_i \geq 0$, and $v_i$ are the various loss functions or objectives.

After performing a test/train split, the model was initially trained for 10,000 trials using hyperparameters set to $\delta = \sqrt{0.15}$, $\epsilon_{ID} = 10^{-4}$, and $\epsilon_{gm} = 10^{-3}$, where $\epsilon_{ID}$ and $\epsilon_{gm}$ correspond to hyperparameter $\epsilon$ for $I_D$ and $g_m$, respectively. The weights were set to $w_1 = 0.50$ and $w_2 = 0.50$, representing the weighted sums of the errors for $I_D$ and $g_m$, respectively. Having established the loss function and its corresponding hyperparameter values, the model was re-trained for an additional 2,500 trials on the entire dataset, starting with the best parameters identified during the train/test split step. The minimum error observed to date is plotted in Fig. 8 as a function of the number of trials, showing the train and test curves, along with the error curve when retraining on the entire dataset. We note that the best set of model parameters was found in under 6,000 trials, resulting in an excellent fit with an error of 1.25e-3 for current values $> 10^{-4}$ A and an error of 2.17e-3 for transconductance values $> 10^{-3}$ S. The measured and simulated $I_D$ and $g_m$ characteristics are shown in Fig. 9.

As in the diode example, we assess robustness by intentionally corrupting 25% of the measurements randomly. In this instance, we introduce outliers by transforming the DC characteristics from the linear domain to the logarithmic domain, corrupting the data, and then reverting it to the linear domain. These outliers follow a normal (Gaussian) distribution with a mean ($\mu$) of 0 and a standard deviation ($\sigma$) of 10. We follow the same procedure as in the diode example, where we use the loss functions from (5) and (6) to test their robustness against measurement outliers. The minimum error observed to date for both loss functions is plotted in Figs. 10(a) and 10(b) as a function of the number of trials. This includes the train and test curves and the error curve when re-training on the entire dataset (for 2,500 trials). The logarithmic and linear representations of the measured and simulated $I_D$ characteristics are displayed in Fig. 11. Once again, this shows that our method is robust against outliers. By using the robust loss function defined in (5), our extracted model achieved an error of 1.27e-3 for currents > $10^{-4}$ A and an error of 2.53e-3 for transconductance values > $10^{-3}$ S when compared against the non-corrupted data. On the other hand, the non-robust loss function, as defined in (6), resulted in a poor fit, with an error of 0.325 for currents > $10^{-4}$ A and an error of 0.238 for transconductance values > $10^{-3}$ S when compared against the non-corrupted data.

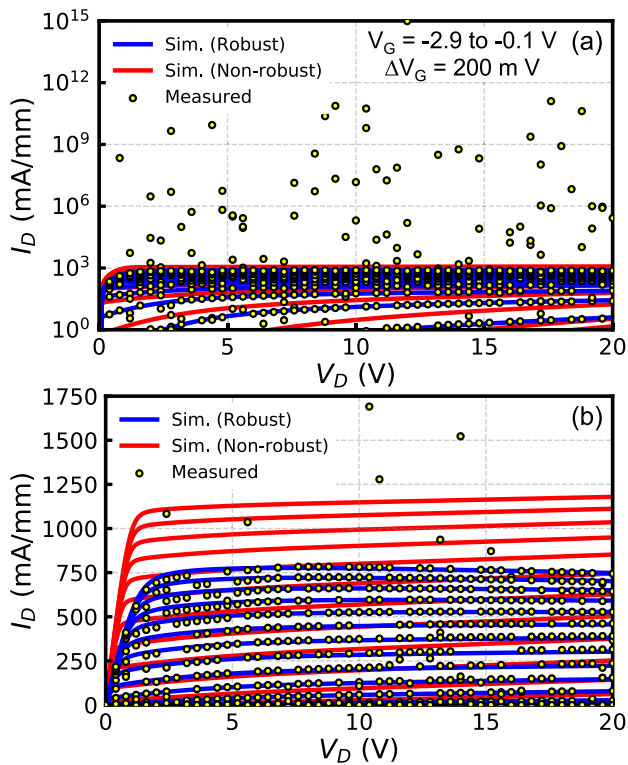

**FIGURE 11.** Measured and simulated $I_D$ characteristics of the GaN HEMT in the (a) log and (b) linear regimes using both robust and non-robust loss functions. The drain voltage $V_D$ was swept from 0 to 20 V with a step size of 100 mV, and the gate voltage $V_G$ was swept from −2.9 to −0.1 V with a step size of 200 mV. Here, 25% of the measurements were randomly corrupted based on a normal distribution ($\mu = 0$ and $\sigma = 10$).

## V. CONCLUSION

In this paper, we have addressed the problem of model parameter extraction via derivative-free optimization. We propose using a loss function that ensures consistent model performance across different orders of magnitude, prioritizes accuracy above a specific measurement or simulation threshold, and remains robust against outliers and measurement errors. To demonstrate how we address these three critical issues, we consider two examples, beginning with the modeling of a diamond Schottky diode. Our focus is then shifted to extracting the ASM-HEMT DC model in a 150-nm GaN-on-SiC HEMT by simultaneously extracting 35 model parameters. Our proposed approach yields good results in a fraction of the simulations required by other approaches in the literature. For instance, the work in [8] demands 120,000 simulation runs for training and testing while considering only 10 ASM-HEMT model parameters. This deep learning approach takes a significant amount of time to gather the required training and testing data and could require an even more substantial number of simulations if additional model parameters are considered. On the other hand, our approach achieved a good fit in a fraction of that amount (i.e., < 5%), demonstrating the usefulness of DFO in device modeling.

The present work opens up several avenues for future research. One potential direction is to explore the parameter extraction of compact models beyond GaN, particularly considering models that require adjusting more than 50 parameters (e.g., BSIM-CMG). Although DFO methods typically become less efficient as the number of parameters increases, the limit of the exact number of parameters in the context of compact model parameter extraction is yet to be established, as this is an emerging area of study. A logical pathway for further investigation would be to explore other derivative-free optimization algorithms beyond TPE and systematically quantify their performance in this context. Comparing the performance of various derivative-free optimization algorithms could provide deeper insights and potentially identify more effective approaches for handling high-dimensional parameter spaces. Another promising pathway within the scope of this research is calibrating TCAD models using our proposed approach. This could be highly beneficial for TCAD engineers, as such calibrations generally require weeks or months due to the long simulation times. Lastly, to foster further exploration of our approach within the device modeling community, we are making our code open-source. It is implemented to support both commercial and fully open-source tools and is available in our GitHub repository.[1]

## ACKNOWLEDGMENT
The authors would like to acknowledge Keysight Technologies Management for supporting this work.

---

[1] https://github.com/rafapm/dfo_parameter_extraction

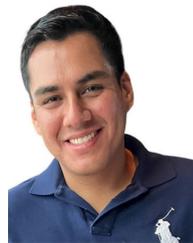

**RAFAEL PEREZ MARTINEZ** (Graduate Student Member, IEEE) received the B.S. degree in electrical engineering and the M.S. degree in electrical and computer engineering from Georgia Institute of Technology, Atlanta, GA, USA, in 2016 and 2019, respectively, and the M.S. degree in electrical engineering from Stanford University, Stanford, CA, USA, in 2023, where he is currently pursuing the Ph.D. degree in electrical engineering.

From Summer 2022 to Summer 2023, he was an Intern with Keysight Technologies Inc., Santa Rosa, CA, where he was responsible for GaN device modeling, characterization, and circuit design. His current research interests include semiconductor modeling, machine learning, optimization, and electronic design.

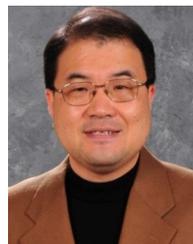

**MASAYA IWAMOTO** (Member, IEEE) received the B.S. degree in electrical engineering from Cornell University, Ithaca, NY, USA, in 1997, and the M.S. and Ph.D. degrees in electrical engineering from the University of California at San Diego, La Jolla, in 1999 and 2003, respectively.

He has been with Keysight Technologies (formerly Agilent Technologies), Santa Rosa, CA, USA, since 2003, where he is involved in the area of GaAs and InP-based IC technology development. He is the key developer of the ADSHBT Model (formerly AgilentHBT Model), which is available in several of Keysight's circuit design software products and has been used in the design of GaAs and InP HBT ICs for Keysight's microwave/mm-wave instrumentation. Currently, he is working on the reliability of HBTs and HEMTs in various III-V material systems (InP, GaAs, and GaN), where his research interest is to apply advanced modeling and characterization techniques in the evaluation of these devices.

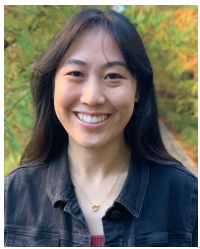

**KELLY WOO** (Graduate Student Member, IEEE) received the B.S. degree in electrical engineering from California Institute of Technology, Pasadena, CA, USA, in 2018, and the M.S. degree in electrical engineering from Stanford University, Stanford, CA, in 2020, where she is currently pursuing the Ph.D. degree in electrical engineering.

Her current research interests include fabricating various high-power electronic devices based on diamond technology and other wide-bandgap materials.

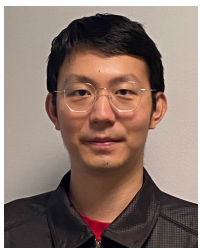

**ZHENGLIANG BIAN** (Graduate Student Member, IEEE) received the B.S. degree in engineering mechanics from Tsinghua University, Beijing, China, in 2019, and the M.S. degree in electrical engineering from Stanford University, Stanford, CA, USA, in 2022, where he is currently pursuing the Ph.D. degree in mechanical engineering.

His research interests include GaN-based high-power and high-frequency devices and novel thermal management methods for these devices.

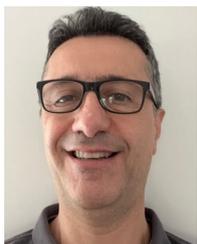

**ROBERTO TINTI** received the Ph.D. degree in electrical engineering from Delft Technical University, The Netherlands, in 1999.

Since joining Keysight Technologies (formerly HP), in 1999, he has been involved in the area of advanced device characterization, including the development of early pulsed and 1/f noise measurement systems. He has contributed significantly to several areas of the IC-CAP device modeling platform, such as optimization and GaN modeling extraction flows. Recently, he became the Product Owner of Keysights device modeling software products. Currently, he is focusing on advancing Keysight EDA design solutions to address current and future challenges in device modeling.

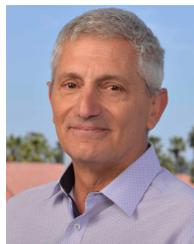

**STEPHEN BOYD** (Life Fellow, IEEE) received the A.B. degree in mathematics from Harvard University, Cambridge, MA, USA, in 1980, and the Ph.D. degree in electrical engineering and computer science from the University of California at Berkeley, CA, USA, in 1985.

He is currently the Samsung Professor in engineering and a Professor in electrical engineering with Stanford University, Stanford, CA. His current research interests include convex optimization applications in control, signal processing, machine learning, and finance. He is a member of the U.S. National Academy of Engineering (NAE) and a Foreign Member of the Chinese Academy of Engineering (CAE) and the National Academy of Engineering of Korea (NAEK).

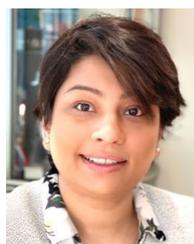

**SRABANTI CHOWDHURY** (Fellow, IEEE) received the M.S. and Ph.D. degrees in electrical and computer engineering from the University of California at Santa Barbara, Santa Barbara, CA, USA, in 2008 and 2010, respectively.

She is an Associate Professor of electrical engineering and a Senior Fellow of the Precourt Institute for Energy and Materials Science and Engineering (by courtesy), Stanford University. She demonstrated the first vertical power-switching transistor in GaN, known as CAVET. She specializes in wideband gap and ultra-wide bandgap materials and device engineering, with a focus on energy-efficient system architecture and thermal management. She received the 2023 Technical Excellence Award from SRC for her work on diamond integration with GaN and SiC, the 2020 Alfred P. Sloan Fellowship in Physics, and the 2016 Young Scientist Award at the International Symposium of Compound Semiconductors (ISCS) for developing Vertical GaN transistors. She actively serves on various IEEE committees, including EDS, VLSI, and IEDM.

· · ·